\documentclass[letterpaper, 10 pt, conference]{ieeeconf}  
\IEEEoverridecommandlockouts                
\usepackage{graphicx}
\usepackage{amssymb}
\usepackage{amsmath}
\usepackage{amsfonts}
\usepackage{multirow}
\usepackage{hyperref}
\usepackage{makecell}
\usepackage{subcaption}
\usepackage{cite}
\usepackage{color}

\title{\LARGE \bf
FD-SLAM: 3-D Reconstruction Using Features and Dense Matching 
}

\author{Xingrui Yang$^{1}$, Yuhang Ming$^{1}$, Zhaopeng Cui$^{2}$ and Andrew Calway$^{1}$%
\thanks{$^{1}$The authors are with the Visual Information Laboratory, Department of Computer Science, University of Bristol, Bristol, U.K.
{\tt\small \{x.yang, yuhang.ming,andrew.calway\}@bristol.ac.uk}}%
\thanks{$^{2}$Zhaopeng Cui is with the State Key Laboratory of CAD\&CG, Zhejiang University, Hangzhou, China.
{\tt\small zhpcui@zju.edu.cn}}%
}

\begin{document}

\maketitle
\thispagestyle{empty}
\pagestyle{empty}

\begin{abstract}
It is well known that visual SLAM systems based on dense matching are locally accurate but are also susceptible to long-term drift and map corruption. In contrast, feature matching methods can achieve greater long-term consistency but can suffer from inaccurate local pose estimation when feature information is sparse. Based on these observations, we propose an RGB-D SLAM system that leverages the advantages of both approaches: using dense frame-to-model odometry to build accurate sub-maps and on-the-fly feature-based matching across sub-maps for global map optimisation. In addition, we incorporate a learning-based loop closure component based on 3-D features which further stabilises map building. We have evaluated the approach on indoor sequences from public datasets, and the results show that it performs on par or better than state-of-the-art systems in terms of map reconstruction quality and pose estimation. The approach can also scale to large scenes where other systems often fail.

\end{abstract}
\section{INTRODUCTION}

Real-time 3-D reconstruction of dense scene models using a handheld RGB-D camera has been actively studied for many years, but still remains an open problem. Most works aim to build consistent and corruption-free 3-D maps with good accuracy and high efficiency. The main obstacle in this area is that dense maps can be easily corrupted as inconsistent observations are made, such as when encountering a loop, large tracking drift, etc. Most works~\cite{newcombe:2011:kinfu, whelan:2015:efusion} use a monolithic 3-D map to represent the scene, which may lead to map corruption and tracking failure. 

Existing approaches to this challenge can be grouped into two categories. The first one is on-the-fly map correction after the camera trajectory gets optimised~\cite{whelan:2015:efusion, maier:2017:correctable, dai:2017:bundlefusion, schops:2019:badslam}. These methods are able to handle small loopy trajectories as they can reduce tracking drift by constantly localising the camera w.r.t. a dense map~\cite{newcombe:2011:kinfu}. However, they either solve a sub-optimal optimisation problem, e.g., pose graph~\cite{maier:2017:correctable} or deformation graph~\cite{whelan:2015:efusion}, or are too computationally demanding as in~\cite{dai:2017:bundlefusion, schops:2019:badslam}.

Another approach is to build sub-maps and integrate them together when needed \cite{maier:2017:correctable, kahler:2015:inftam, reijgwart:2020:voxgraph}. However, large sub-maps tend to create registration artefacts around the boundaries, since sub-maps can only be rigidly transformed~\cite{kahler:2015:inftam}. Moreover, creating sub-maps too often~\cite{maier:2017:correctable} can lead to increasing drifts between maps, especially for loopy motion. Although some can handle long-term loop constraints~\cite{reijgwart:2020:voxgraph}, they still fail to handle tracking lost due to the lack of explicit sub-map retrieval and alignment.

\begin{figure}[t]
\centering
  \subcaptionbox{Ours}{\includegraphics[width=.5\columnwidth]{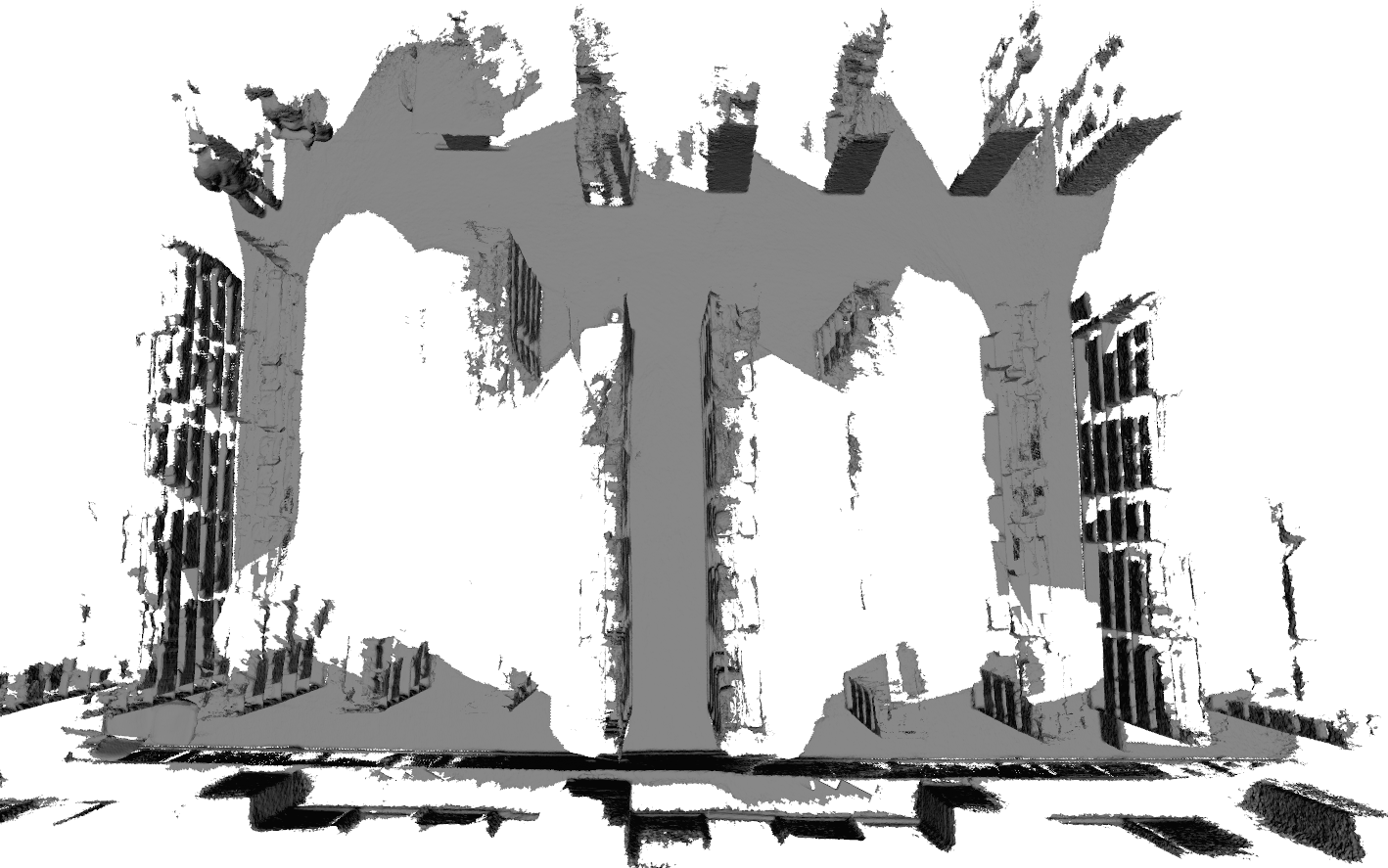}}%
  \subcaptionbox{InifiTAM}{\includegraphics[width=.5\columnwidth]{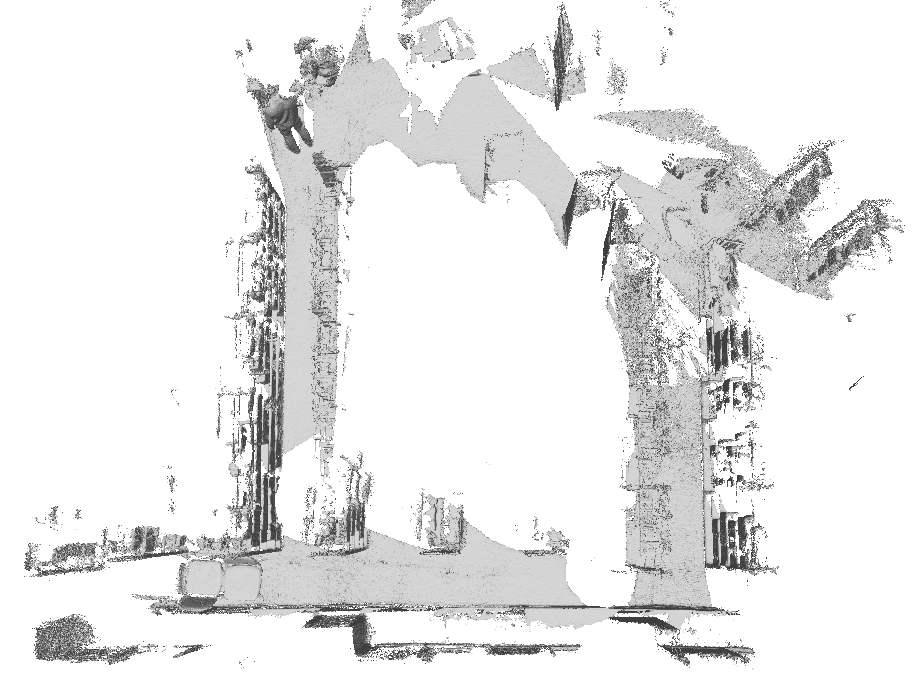}}
  \subcaptionbox{ElasticFusion}{\includegraphics[width=.5\columnwidth]{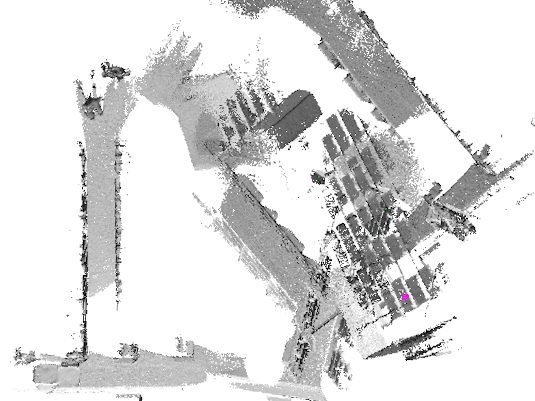}}%
  \subcaptionbox{An example image}{\includegraphics[width=.5\columnwidth]{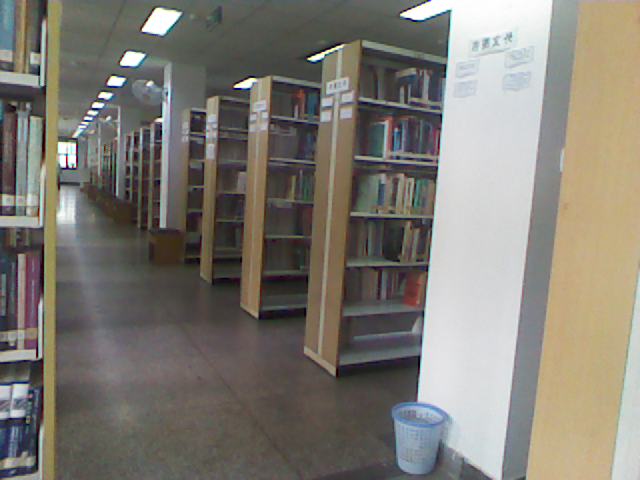}}
\caption{Reconstruction of a large-scale library floor~\cite{liu:2019:npu} from (a) our system (b) InfiniTAM~\cite{kahler:2015:inftam} and (c) ElasticFusion~\cite{whelan:2015:efusion}. An example frame is shown in (d). This is a challenging scene for dense SLAM systems as it contains repetitive features and shaky camera motions. Both ElasticFusion and InfiniTAM completely failed on this scene. While our reconstruction is consistent and corruption free, owing to our efficient feature-based back-end.}
\label{fig:demo}
\vspace{-2em}
\end{figure}

To address these problems, we present a new RGB-D SLAM system, which exploits the advantages of both dense and feature-based matching. Specifically, our system uses a frame-to-model visual odometry to register incoming RGB-D frames into volumetric sub-maps. Simultaneously, motivated by feature-based systems such as ORB-SLAM3~\cite{campos:2021:orb3}, we maintain a globally consistent sparse feature map. We build this sparse map by detecting and matching ORB features from existing keyframes. This sparse global map allows us to build constraints between 3-D points and sub-maps in addition to the pairwise sub-map constraints, which benefits the global pose optimisation. Lastly, we incorporate a fast sub-map retrieval and matching framework based on learned 3-D features to handle loop detection and relocalisation efficiently, helping to reduce long-term drift error.

To summarise, our key contributions are: (1) We present a novel large-scale dense RGB-D SLAM pipeline, which achieves real-time indoor mapping by combining a dense frame-to-model odometry with a feature-based back-end; (2) We propose a new sub-map retrieval and alignment method to effectively handle long-term loop correction and relocalisation; (3) Both quantitative and qualitative experiments show the proposed method has achieved on par performance with state-of-the-art systems while surpassing them on challenging large-scale scenes. 

\section{Related Work}

The advent of inexpensive RGB-D cameras has made possible real-time dense reconstruction without planned sensor trajectories. Early works took inspiration from traditional multi-view systems and use feature points to localise the camera~\cite{stein:2014:cpufusion}. The depth camera is only used as a way to initialise 3-D features. The seminal work of KinectFusion~\cite{newcombe:2011:kinfu} introduced a voxel map representation~\cite{curless:1996:sdf} and a frame-to-model tracking strategy which leads to significant improvements on workshop-scale reconstruction. 

\cite{roth:2012:mkinfu, whelan:2015:largefusion} proposed a shifting volume approach to expand KinectFusion to large areas. However, a lot of memory is still wasted on reconstructing empty space. \cite{meillan:2013:unify} represent the map as keyframes, and novel views can be directly rendered from nearby keyframes. \cite{stein:2014:cpufusion, hornung:2013:octomap, vespa:2018:octreeslam} build volumetric map using an octree that only reconstructs occupied space. This representation is suitable for motion planning. \cite{dai:2017:bundlefusion, kahler:2015:inftam, niebner:2013:voxelhashing} use a hash table to store voxels which allows fast access and modification. 

ElasticFusion~\cite{whelan:2015:efusion} uses surface elements (surfels) instead of voxels to represent the map, which allows the map to be changed on-the-fly. They use a deformation graph to deform the map non-rigidly when a loop is found. \cite{schops:2019:badslam} also uses a surfel map and performs full bundle adjustment to jointly optimise camera pose and map structure. In comparison, our work is focused on the efficient optimisation of voxel maps.  

To solve the map corruption problem for voxel maps, BundleFusion~\cite{dai:2017:bundlefusion} introduced an on-the-fly map correction algorithm that re-registers each frame after their poses get optimised. However, this is computationally demanding and does not scale well to large-scale scenes. To increase the efficiency of this process, \cite{maier:2017:correctable} proposed to fuse frames into their reference keyframes before applying map correction. However because a large chunk of data outside of the keyframes' field of view is discarded, their method will lead to incomplete reconstructions.

Closely related to our works, many researchers have studied to build sub-maps~\cite{kahler:2015:inftam, reijgwart:2020:voxgraph, henry:2013:patchvolume, fioraio:2015:subvolume}. Most works~\cite{kahler:2015:inftam, reijgwart:2020:voxgraph} use pose graph to optimise inter-map relations, which leads to visible seams between maps. \cite{reijgwart:2020:voxgraph , fioraio:2015:subvolume} optimises the map by explicitly aligning sub-maps and optimising through their TSDF fields. In comparison, we optimise our map by performing a full bundle adjustment and use learned 3-D features to detect long-term loops and localise a lost camera. 

Relocalisation and loop detection are crucial for achieving robust and corruption-free reconstruction, but were ignored by many dense SLAM systems. \cite{whelan:2015:efusion, kahler:2015:inftam} uses random ferns~\cite{glocker:2013:fern} to retrieve similar keyframes and directly match them with iterative closest points (ICP). \cite{li:2015:rgbdreloc} introduced 2-D features into the dense map, and use them to quickly relocalise a lost camera. However, they did not address the problem of loop closure. \cite{reijgwart:2020:voxgraph} introduced loop constraints through the use of bag of visual words~\cite{raul:2014:bow}, but the problem of relocalisation is not discussed.

\section{System Overview}

\begin{figure}[t]
\centering
\includegraphics[width=\linewidth]{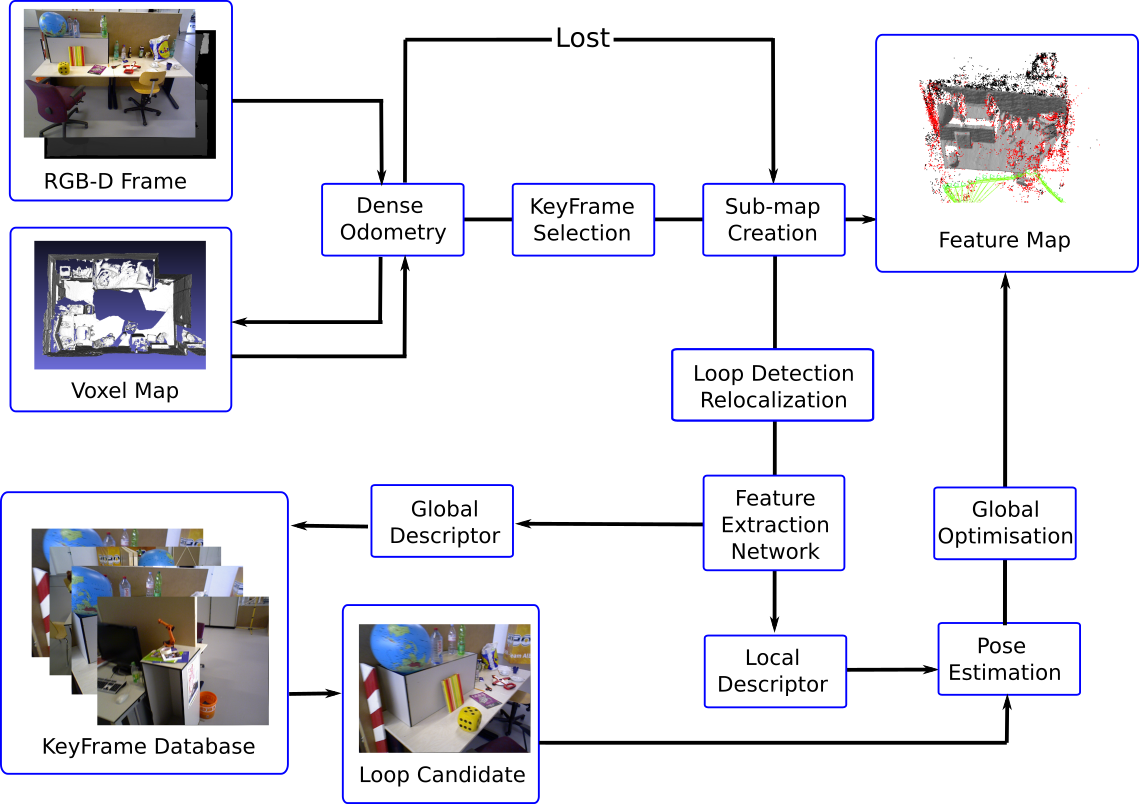}
\caption{Overview of our proposed system. We update the active sub-map when a new frame is successfully localised by the dense odometry. We create new keyframes based on angular and translational changes. For each keyframe we also create a new sub-map. We then extract 3-D features from sub-maps with a neural network. The extracted 3-D features are used to quickly retrieve sub-maps to detect long-term loops and to relocalise a lost camera frame.}
\label{fig:flow}
\vspace{-1em}
\end{figure}

The architecture of our system is illustrated in~\autoref{fig:flow}. As new RGB-D frames are captured by an RGB-D camera (such as Microsoft Kinect, etc.), we estimate their initial poses with a coarse-to-fine, frame-to-model dense odometry. For each frame, we register them into a dense volumetric map.

Our system generates new sub-maps based on the rotational and translational changes. If the changes with the previous reference keyframe are larger than a predefined threshold, we create a new sub-map. For each sub-map there is also a keyframe associated with it. Subsequent tracking and mapping are then performed on this new sub-map. 
In order to achieve long-term map consistency and relocalise lost frames, we employ learned 3-D features for loop detection and relocalisation. We extract local and global features from point clouds generated from each sub-map with a neural network building on~\cite{du:2020:dh3d}. These features provide the ability to match between sub-maps and detect long-term loops. Finally, we use bundle adjustment~\cite{triggs:2000:ba} to jointly optimise the position of feature points and the pose of sub-maps. We will elaborate on each part of our system in the following sections.

\section{Dense RGB-D Odometry}

We define the image domain as $\Omega \in \mathbb{N}^2$. Therefore, input images can be represented as $I(\Omega)\in \mathbb{R}^3$ and $D(\Omega)\in\mathbb{R}$. The camera intrinsics $K\in\mathbb{R}^{3\times 3}$ is also given a priori. We aim to estimate a 3-D transformation $T\in SE(3)$ for each frame in real-time, which are then used for subsequent optimisations. 

We employ an iterative optimisation process similar to the tracking method used in~\cite{whelan:2015:efusion} to align input frames to the active sub-map. Specifically, We are aiming to find a 3-D transformation $T^{*}$ that best aligns the frame to the map. This is done by jointly optimising a geometric error term and a photometric error term:

\begin{equation}
r_g(\xi)=\mathbf{n_i}\cdot(\mathbf{p_i}-\hat{\xi}\mathbf{T_i}\mathbf{q_j}),
\end{equation}
\begin{equation}
r_I(\xi)=I_1(\Omega)-I_2(K(\hat{\xi}\mathbf{T_i} K^{-1}(\Omega,D(\Omega)))),
\end{equation}

where $T_i$ is the current camera pose, $\xi\in \mathfrak{se}(3)$ is the incremental changes in the tangent space, which we aim to estimate, $\mathbf{p_i}$ and $\mathbf{q_j}$ are corresponding points from the source and reference frames respectively, and $\mathbf{n_i}$ is the corresponding surface normal sampled at the source frame. We jointly optimise these two terms iteratively using iteratively re-weighted least squares. We also adopt Huber norm and weight each residual by its inverse depth, since the error from RGB-D cameras is inversely proportionate to the depth measurement.

\section{Sub-map Generation}

\begin{figure}[t]
\centering
\includegraphics[width=\linewidth]{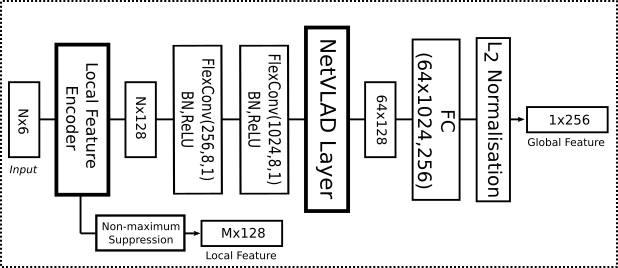}
\caption{An illustrative sketch of our feature extraction network. We take $N$ 3-D points $p=(x,y,z)$ as input, and each point is associated a colour $c=(r,g,b)$. We extract from the set of points $M$ local features of shape $1\times 128$ and a global descriptor of shape $1\times 256$.}

\label{fig:network}
\vspace{-1.5em}
\end{figure}

Our system generates a new sub-map whenever the pose differences between the reference sub-map and the current camera frame become larger than a threshold. Each sub-map $M_i$ is represented by 3-D voxels. Each voxel stores a truncated signed distance (TSDF) $s^i\in [-1,1]$, with the value indicating the distance of the voxel to the nearest surface. Hence the surface is implicitly parameterised as the zero-level set of this scalar field. Each voxel also has colour $c^i\in \mathcal{N}^3$ and weight $w^i\in \mathcal{N}$ attached to it. 

Once the pose of an RGB-D frame has been estimated, we fuse the frame into the current sub-map. We update voxels the same way as in~\cite{newcombe:2011:kinfu}. Although different weighting algorithms have been proposed to account for sensor uncertainties~\cite{oleynikova:2017:voxblox}, we find this simple scheme shows no degradation in the quality of reconstructions since our voxels are sufficiently small. We use a hash table to store and index voxel blocks, which allows $\mathcal{O}(1)$ time insertion and searching, similar to methods used in~\cite{niebner:2013:voxelhashing, kahler:2015:inftam}.

To find the zero-level set of the reconstructed map, we cast rays from the camera centre through each of the pixels $(x,y) \in \Omega$ on the image plane and find their intersections with the reconstructed surface. Each time the map is updated, we also update the stored surface points for the reference frame. This method is called frame-to-model tracking and is known to improve localisation accuracy for dense SLAM~\cite{newcombe:2011:kinfu}.

\section{Sparse Feature-based Map}

Different from other sub-map based systems that rely on pose graph of pairwise constraints. We extract sparse 3-D points that are associated with 2-D feature points from each sub-map to build a feature point map. More specifically, we detect ORB~\cite{rublee:2011:orb} features from the keyframes of each sub-map, and find their corresponding depth from the map through a ray casting process.

In feature-based systems, 2-D features are usually matched by finding their nearest neighbours in the descriptor space, which is inefficient and prone to outliers. Instead, we use a projective feature association method, similar to what was used in~\cite{hsiao:2017:planarslam}, to quickly find 3D-2D correspondences. 

To find the corresponding points, we project 3-D points that are observed from nearby keyframes to the current keyframe. Then we search for correspondences within a radius around the centre of the projection. A match is found when all of the following criteria are met: (1) The searched feature has valid depth; (2) The distance between both points is within a threshold $\delta p$; (3) The ORB descriptor distance is smaller than a threshold. We further optimise the camera pose based on the reprojection error of matched features. The potential outliers are pruned during this step. 

\section{Relocalisation and Loop Detection}

\begin{figure*}[!t]
\rotatebox[origin=t]{90}{Ground Truth} \quad
\begin{subfigure}{0.95\linewidth}
\centering
\includegraphics[width=\linewidth]{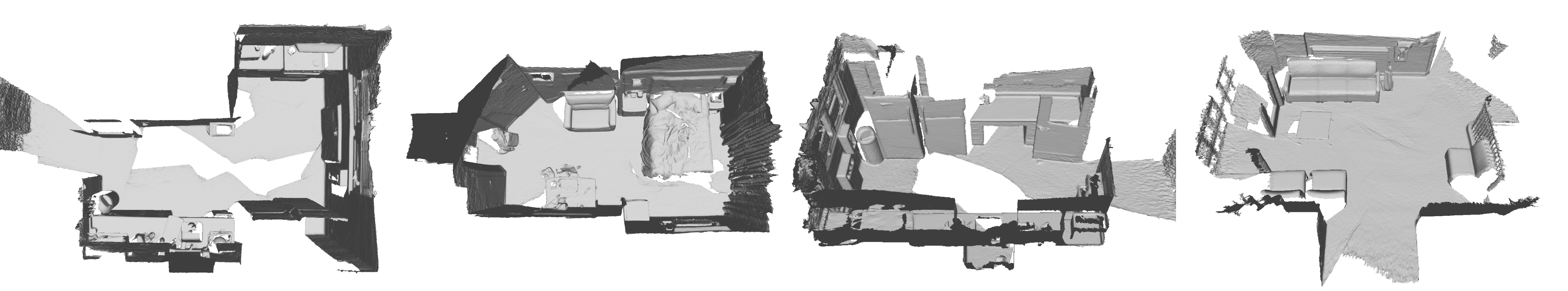}
\end{subfigure}%

\rotatebox[origin=t]{90}{Ours} \quad
\begin{subfigure}{0.95\linewidth}
\centering
\includegraphics[width=\linewidth]{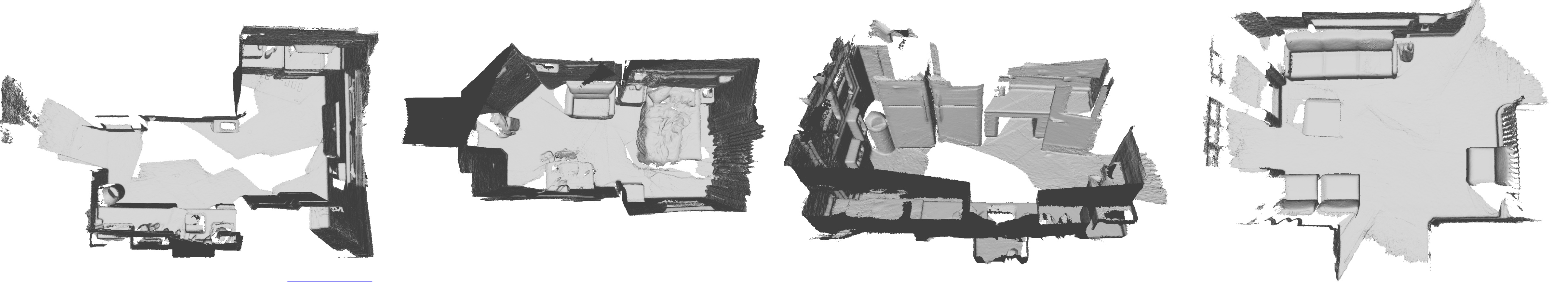}
\end{subfigure}%

\rotatebox[origin=t]{90}{ElasticFusion} \quad
\begin{subfigure}{0.95\linewidth}
\centering
\includegraphics[width=\linewidth]{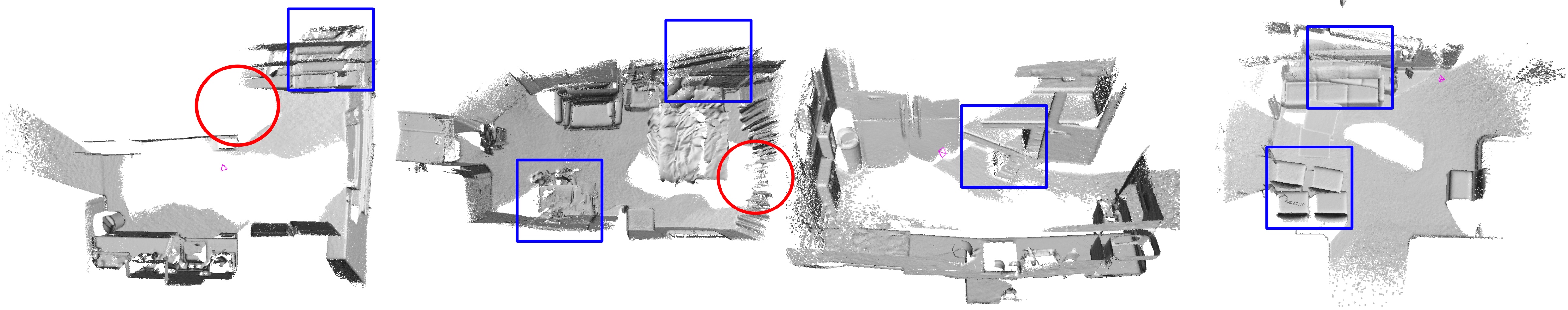}
\end{subfigure}
\caption{From top to bottom: reconstruction obtained from ground truth poses, our system and ElasticFusion. The reconstruction from ElasticFusion is inconsistent because of failure to detect loops (blue squares) and missing data (red circle).}
\label{fig:recon_qual}
\vspace{-1.5em}
\end{figure*}

To deal with challenging indoor scenes, which often leads to tracking lost and large drifts, we propose an effective sub-map retrieval and alignment method for loop detection and relocalisation. 

Motivated by~\cite{du:2020:dh3d}, we train a neural network to directly extract 3-D feature points from coloured point clouds. The architecture of our network is displayed in~\autoref{fig:network}: The input point cloud is first processed by a local feature extraction network, which consists of a series of $1\times 1$ convolution with batch normalisation. The local network produces $M=512$ salient features. These local features are subsequently processed by multiple flex-convolution~\cite{groh:2018:flexconv} layers and finally aggregated by a NetVLAD layer~\cite{uy:2018:pointnetvlad} to form a global descriptor.

Different from relevant works that only use 3-D coordinates of the points \cite{du:2020:dh3d, uy:2018:pointnetvlad}, the input to our network is a set of coloured points, each point contains a 3-D coordinate and RGB colour. The included colour allows our system to disambiguate geometrically similar scenes.

\subsection{Network Training}

We generate our training data from ScanNet \cite{dai:2020:scannet}, which is a large indoor RGB-D dataset. To prepare data, we first fuse every 50 frames into a dense map for each train sequence. We then generate a point cloud by querying the map for voxels that has a TSDF value smaller than a threshold, similar to~\cite{zeng:2017:3dmatch}. Unlike most outdoor datasets \cite{du:2020:dh3d} where GPS signals can be used to determine matching scans for descriptor learning, we use ground-truth poses provided in the dataset to test if two scans belong to the same scene. The method we use to check co-visibility is further elaborated in~\autoref{sec:FR}. 

Following DH3D~\cite{du:2020:dh3d}, we train our network in two separate steps: We start by training the local feature extraction network supervised by the $N$-tuple loss~\cite{deng:2018:ppfnet}. We select $6$ pairs of point clouds for every batch. Then we fix the weights of the local network and train the global one with the lazy quadruplet loss as described in~\cite{uy:2018:pointnetvlad}. Each iteration we randomly choose an anchor point cloud with 2 positives and 4 negatives. Each step is trained for 20 epochs. We also augment the training data with random Gaussian noise and rotations.

\subsection{Loop Detection}

We detect loop candidates when a new sub-map is created, similar to~\cite{whelan:2015:efusion, campos:2021:orb3}. For each sub-map created, we extract from it a set of salient points, associated with local feature descriptors and a global descriptor. The sub-map is converted to a point cloud by finding the zero-level set of the TSDF volume, and colours are interpolated from the input images based on the found points. To check candidate sub-maps, we find $k$ nearest neighbours of the new sub-map from the global descriptor space. 

We try to align candidate sub-maps with the query sub-map using their local descriptors. The relative poses are estimated by applying a RANSAC-based absolute orientation algorithm~\cite{Horn:1988:ao}. Once a sub-map is found with a sufficient number of matching features, we double-check the feature correspondences by running an additional dense alignment step to ensure accurate alignment.

\subsection{Relocalisation}

Our sub-map retrieval and alignment scheme enables on-the-fly relocalisation of lost cameras. Since only the current sub-map is used by the odometry, whenever tracking fails, we simply create a new sub-map and continue localising the camera with it. Then we keep trying to relocalise the new sub-map w.r.t. the previously built maps. The same network that we use to detect loops is also used here to match sub-maps. Once a match is found, we fuse matching feature points and matching sub-maps similar to \cite{campos:2021:orb3}.

\section{Sub-map Fusion}

To provide a global 3-D map for downstream tasks, such as semantic segmentation, our system is capable of combining all sub-maps together by merging voxels from adjacent maps. Since all sub-maps are axis-aligned to their respective local coordinate, we use a sampling-based strategy to achieve accurate map fusion. This is done in three steps: (1) For each voxel in the target map, we create a corresponding voxel in the host map if it does not exist; (2) For each voxel in the host map, we warp its positions to the target map, and find its TSDF value, colour and weight via interpolation; (3) The new values are then combined with the old ones in the same way as described in the previous section. 

Although the map fusion process is fast for small scenes due to parallel GPU processing, it can be costly for large-scale maps since the processing time scales linearly against the number of sub-maps in the system. 

\section{Evaluation}

\subsection{Reconstruction}

\begin{table}[!t]
\caption{Reconstruction Results on Small Dataset}
\label{tab:recon}
\begin{center}
\begin{tabular}{|c||c||c||c||c|}
\hline
Seq. & Metric & \makecell{Our \\System} & \makecell{Elastic \\Fusion} & \makecell{DVO \\SLAM} \\
\hline
\multirow{2}{*}{kt0} & ATE & \textbf{0.007} & 0.008 & 0.102 \\ 
& Mean Dst. & \textbf{0.004} & 0.007 & 0.032 \\
\hline
\multirow{2}{*}{kt1} & ATE & 0.012 & \textbf{0.010} & 0.031 \\ 
& Mean Dst. & 0.009 & \textbf{0.007} & 0.061 \\
\hline
\multirow{2}{*}{kt2} & ATE & 0.017 & \textbf{0.015} & 0.192 \\ 
& Mean Dst. & 0.009 & \textbf{0.008} & 0.119 \\
\hline

\end{tabular}
\end{center}
\vspace{-2em}
\end{table}

To measure the reconstruction quality, we first tested our system on a small synthetic RGB-D dataset~\cite{handa:2014:iclnuim}. The results are listed in~\autoref{tab:recon}. We compared our system with other commonly-used dense SLAM systems including DVO-SLAM~\cite{kerl:2013:dvo} and ElasticFusion~\cite{whelan:2015:efusion}. Since DVO-SLAM does not reconstruct dense maps, we use their estimated camera poses to create a TSDF map offline. We measure pose accuracy with absolute trajectory error (ATE) as the same with~\cite{sturm:2012:tumrgbd}. We also measure reconstruction quality as the mean distance of reconstructed surface points to the ground truth model using the script provided by~\cite{handa:2014:iclnuim}. These results show that our sub-map approach performs on par with ElasticFusion on reconstructing small scenes. We also show better performance against DVO-SLAM on all scenes. 

We also compared our system with ElasticFusion on the testing set of ScanNet. Since ElasticFusion failed on most of the scenes, We only show some qualitative results in~\autoref{fig:recon_qual}. The comparison of camera trajectory estimation is provided in the next section. 
As can be seen, our system correctly recognises large loops, resulting in a consistent and clean map, while the results from ElasticFusion have open loops. ElasticFusion also creates incomplete reconstructions, since surfaces that are only observed a few times cannot be correctly initialised in their system, whereas our voxel-based system can correctly reconstruct surfaces that are even observed only once, which results in a more complete map representation.

\begin{table}[!t]
\caption{Trajectory Estimation Results on Scannet Dataset}
\label{tab:pose}
\begin{center} 
\begin{tabular}{|c||c||c||c||c|}
\hline
Seq. & Metric & Ours & \makecell{Elastic \\Fusion\cite{whelan:2015:efusion}} & \makecell{ORB \\SLAM3\cite{campos:2021:orb3}} \\
\hline
\multirow{2}{*}{707} & ATE & \textbf{0.0693} & 0.1319 & 0.0763 \\ 
& RPE & \textbf{0.0890} & 0.1660 & 0.0960 \\
\hline
\multirow{2}{*}{709} & ATE & \textbf{0.0417} & 0.0554 & 0.0623 \\
& RPE & \textbf{0.0587} & 0.0746 & 0.0854 \\
\hline
\multirow{2}{*}{714} & ATE & 0.0560 & 0.1690 &  \textbf{0.0550}  \\
& RPE & \textbf{0.0784} & 0.1870 & 0.0804 \\
\hline
\multirow{2}{*}{719} & ATE & 0.0292 & 0.0299 & \textbf{0.0278} \\
& RPE & 0.0513 & \textbf{0.0440} & 0.0478 \\
\hline
\multirow{2}{*}{722} & ATE & \textbf{0.0326} & 0.0410 & 0.0408 \\
& RPE & \textbf{0.0464} & 0.0623 & 0.0559 \\
\hline
\multirow{2}{*}{738} & ATE & 0.1638 & 0.0985 & \textbf{0.0677}\\
& RPE & 0.2000 & 0.1203 & \textbf{0.0935} \\
\hline
\multirow{2}{*}{741} & ATE & 0.0981 & 0.0683 & \textbf{0.0669} \\
& RPE & 0.0980 & 0.0829 & \textbf{0.0822} \\
\hline
\multirow{2}{*}{746} & ATE & 0.1175 & 0.2324 & \textbf{0.0801} \\
& RPE & 0.2219 & 0.2815 & \textbf{0.1074} \\
\hline
\multirow{2}{*}{748} & ATE & \textbf{0.0487} & 0.0519 & 0.1691 \\
& RPE & \textbf{0.0786} & 0.0821 & 0.2350 \\
\hline
\multirow{2}{*}{760} & ATE & 0.0716 & \textbf{0.0635} & 0.0797 \\
& RPE & \textbf{0.0928} & 0.0995 & 0.1044 \\
\hline
\multirow{2}{*}{782} & ATE & 0.1040 & 1.0274 & \textbf{0.0801} \\
& RPE & 0.1379 & 1.1297 & \textbf{0.1138} \\
\hline
\end{tabular}
\end{center}
\vspace{-2em}
\end{table}

We also tested our system on reconstructing large-scale scenes. However, as this type of datasets usually does not come with ground truth, we can only compare them qualitatively. One such example is displayed in~\autoref{fig:demo}. Most dense reconstruction systems are quite fragile when working with large-scale scenes, they tend to fail quickly when encountering long-term loops, while our system has the ability to correct the loops without worrying about map corruption.

Our system works at a steady frame rate, even with the existence of large loops. This is because the optimisation of the surface reconstruction is done indirectly by adjusting the set of sparse points and sub-maps in our system, which can be efficiently done in a separate thread. In comparison, ElasticFusion maintains a monolithic map and has a frame time proportionate to the number of elements in the map. In addition to that, when a loop is identified, a large portion of the map must be adjusted on-the-fly, which is detrimental to real-time applications. 

\subsection{Trajectory Estimation}

Although our focus is on large-scale reconstruction, it is also very important to make sure that our system can estimate accurate camera poses. We tested our system against ORB-SLAM3~\cite{campos:2021:orb3} and ElasticFusion~\cite{whelan:2015:efusion} on the testing set of ScanNet. Note that ORB-SLAM3 does not have the ability to reconstruct a dense map on-the-fly, it is focused on improving trajectory estimation. The results are shown in~\autoref{tab:pose}. We report the RMSE of absolute trajectory error (ATE) and relative pose error (RPE), as defined in~\cite{sturm:2012:tumrgbd}. All metrics are in meters, and the results are listed in Table~\autoref{tab:pose}. 

As we can see from the table, in terms of ATE, our system performs on par with ORB-SLAM3 and out-performs ElasticFusion on most scenes. Moreover, we show significant improvements over ORB-SLAM3 on RPE over several sequences, which means our dense frame-to-model odometry can deliver accurate relative pose estimates compared to feature-based methods. We also note that ElasticFusion shows lower RPE on a few sequences, as a single dense map can be very effective in certain scenes. Despite also tracking a dense map, our system creates sub-maps constantly, which leads to a slight loss in accuracy. 

It is worth noting that our system is more reliable in most scenes while ElasticFusion suffers occasional break downs due to map corruption. This is an apparent downside of using a monolithic map. Finally, owing to the sub-optimal nature of deformation graph, they have an overall higher absolute error in most scenes, compared to both our system and ORB-SLAM3.

\subsection{Loop Detection and Relocalisation}
\label{sec:FR}

\begin{table}[!t]
\caption{Evaluation on Relocalisation and Loop Detection}
\label{tab:rt1}
\begin{center}
\begin{tabular}{|c||c||c||c||c|}
\hline
sequence & metric & \makecell{ORB\\(RANSAC)} & DH3D \cite{du:2020:dh3d} & Ours \\
\hline
\multirow{3}{*}{\makecell{Textureless Scenes \\(10 sequences)}} & ATE & 0.7693 & 2.45 & \textbf{0.5938} \\
 & ARE & 5.302 & 13.54 & \textbf{3.878} \\
 & SR & 45.29\% & 6.972\% & \textbf{49.2\%} \\
\hline
\multirow{3}{*}{\makecell{Textured Scenes \\(12 sequences)}} & ATE & \textbf{0.7053} & 2.241 & 0.7505\\
 & ARE & \textbf{4.627} & 13.07 & 4.951 \\
 & SR & \textbf{49.57\%} & 10.1\% &42.72\% \\
\hline
\multirow{3}{*}{\makecell{Overall \\(22 sequences)}} & ATE & 0.7522 & 2.367 & \textbf{0.662} \\
 & ARE & 5.099 & 13.4 & \textbf{4.385} \\
 & SR & \textbf{46.67\%} & 8.519\% & 45.44\% \\
\hline
\end{tabular}
\end{center}
\vspace{-2em}
\end{table}

Directly comparing loop detection and relocalisation on different SLAM systems is known to be difficult: Most systems have their detection and relocalisation subroutines tied to their map building process. It is difficult to compare them on a real world setting. Instead, we evaluated the effectiveness of our system on the task of pose estimation between two frames. Many SLAM systems depend on matching two keyframes to correctly relocalise a lost camera or to register loop candidates. Therefore, we have compared our method against DH3D \cite{du:2020:dh3d}, which is also our baseline, and a RANSAC-based pose estimation method using ORB features (ORB+RANSAC)~\cite{Horn:1988:ao}.

We use the official test split of ScanNet in our experiment. As the differences between adjacent frames are often small, we sample consecutive frames such that their rotational difference $\delta R$ and translational difference $\delta t$ with the previously selected frames is larger than a threshold. Here we use $\delta R=0.2rad$ and $\delta t=0.2m$. We found this method generates evenly spaced frames compared to sampling at a fixed time interval as the camera did not move at a fixed velocity.
We then group frames that share enough view. This is done by a frustum overlapping test: Two frames are considered to share a common view if a certain number of points observed in one frame is also observed in the other. 

To determine the number of co-visible points, we first warp points from one frame (source frame) to the other one (reference frame) with their relative transformation, and project the warped points to the image plane of the reference frame. To resolve conflicts when two points are projected to the same position, we employ a depth buffer to only keep the smallest value. The point is said to be co-visible if the distance between its depth after warping and depth value from the projection is within a range. If the number of visible points $N$ is within a threshold, the two frames are said to share a common view. I.e., we choose $N=\alpha M$, where $M$ is the total number of points in the reference frame and $\alpha$ is set to 0.4 in our tests. 

When testing, we randomly generate frame pairs that share common view from the testing set. We calculate the transformation with features from our method, DH3D \cite{du:2020:dh3d} and ORB. The results are displayed as the RMSE of absolute translational error (ATE) and absolute rotational error (ARE). We also define a success rate (SR), which is the percentage of successful matches per category. A match is only successful when both the angular difference $\delta R<0.2 rad$ and translational difference $\delta t<0.2m$. For DH3D, we use the model from their official release. We implemented the RANSAC pose estimation scheme using OpenCV~\cite{bradski:2000:opencv}. The test results on all 22 test scenes are given in~\autoref{tab:rt1}.

\begin{figure}[t]
\centering
\includegraphics[width=\linewidth]{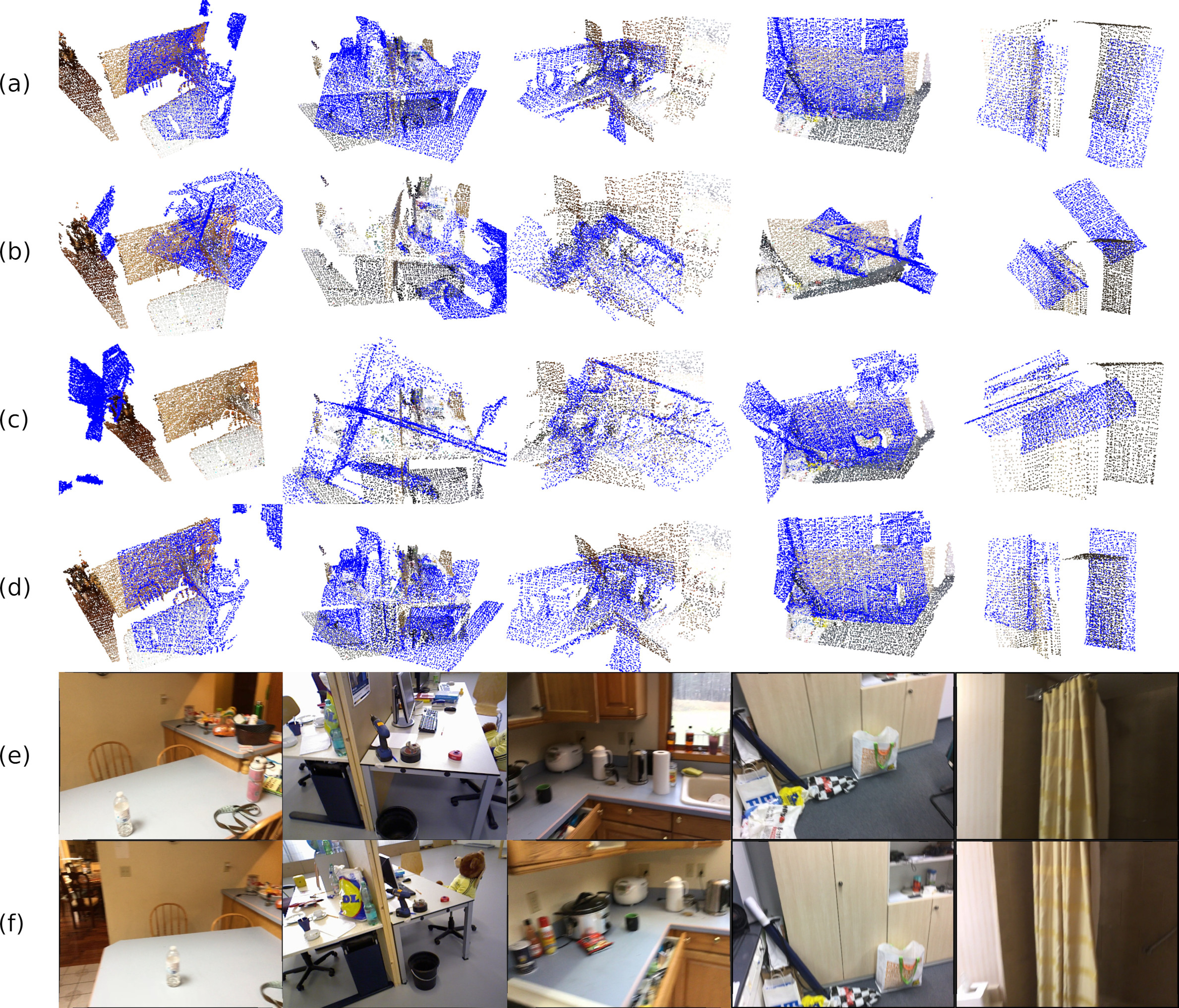}
\caption{Qualitative comparison of local feature matching with other methods (best viewed in colour). We compared (d) our method with (b) DH3D \cite{du:2020:dh3d} and (c) ORB+RANSAC. The ground truth is shown in (a). We put reference frames in their original colour and query frames are coloured in blue for better visualisation. We also show (e) reference frames and (f) query frames. We observed that ORB+RANSAC often struggles with large viewpoint changes and textureless scenes, while DH3D failed on scenes with large dominant planes due to its focus on structural features.}
\label{fig:local_feat_qual}
\vspace{-1em}
\end{figure}

These scenes are classified into two categories based on their appearance, i.e. textureless or textured. The results show our method has a huge improvement compared to DH3D as we also take colour into consideration. Our method also outperformed appearance-based methods on textureless scenes as expected. The experiments also show simple ORB+RANSAC method works well in various indoor scenes, and has an overall higher success rate. However, our method obtained smaller rotational and translational errors, which means our methods are more consistent across the entire test set. We also show some qualitative evaluations in~\autoref{fig:local_feat_qual}. Our method shows clear advantage on scenes with large dominant planes (such as the table in the left most column) and cluttered scenes (e.g. the third column) better than other methods. 

\section{Conclusion}

We present in our paper a real-time dense SLAM system, which takes RGB-D images as input to produce a detailed volumetric map. Our system combines a dense odometry with a sparse feature matching backend to enable accurate localisation. We also exploits learned 3-D features for registering long-term loops and lost cameras. Experiments demonstrate that our system is able to efficiently generate high-quality volumetric reconstructions comparable to other state-of-the-art methods, with advantages in building large-scale maps.

\bibliographystyle{ieeetr}
\bibliography{biblio}

\end{document}